\documentclass[runningheads]{llncs}
\usepackage{graphicx} 
\usepackage{hyperref}
\usepackage{soul}
\usepackage{float}
\usepackage{tcolorbox}
\usepackage{amsmath}
\usepackage{array} 
\usepackage{enumitem} 
\usepackage{tabularx} 

\title{Autoencoder-Based Framework to Capture Vocabulary Quality in NLP}

\author{Vu Minh Hoang Dang\orcidID{0009-0001-0505-314X} \and Rakesh M. Verma\orcidID{0000-0002-7466-7823}}

\titlerunning{Autoencoder-Based Framework to Capture Vocabulary Quality in NLP}
\authorrunning{VMH. Dang and R. M. Verma}

\institute{
University of Houston, USA\\
\email{vdang9@uh.edu},
\email{rmverma2@central.uh.edu}}

\begin{document}

\maketitle

\begin{abstract}
Linguistic richness is essential for advancing natural language processing (NLP), as dataset characteristics often directly influence model performance. However, traditional metrics such as Type-Token Ratio (TTR), Vocabulary Diversity (VOCD), and Measure of Lexical Text Diversity (MTLD) do not adequately capture contextual relationships, semantic richness, and structural complexity. In this paper, we introduce an autoencoder-based framework that uses neural network capacity as a proxy for vocabulary richness, diversity, and complexity, enabling a dynamic assessment of the interplay between vocabulary size, sentence structure, and contextual depth. We validate our approach on two distinct datasets: the DIFrauD dataset, which spans multiple domains of deceptive and fraudulent text, and the Project Gutenberg dataset, representing diverse languages, genres, and historical periods. Experimental results highlight the robustness and adaptability of our method, offering practical guidance for dataset curation and NLP model design. By enhancing traditional vocabulary evaluation, our work fosters the development of more context-aware, linguistically adaptive NLP systems.

\keywords{Lexical Diversity \and Vocabulary Richness Evaluation \and Vocabulary Complexity\and Autoencoder Framework \and Language Evolution}
\end{abstract}

\section{Introduction}

The growing prevalence of natural language processing (NLP) applications has heightened the need for effective measures to evaluate the quality of linguistic datasets. Linguists, however, have been interested in vocabulary measurement for over 50 years, exploring ways to capture lexical richness and diversity across different contexts. Their lexical metrics such as TTR (type-token ratio)~\cite{ttr_yule1944}, MTLD (Measure of Lexical Text Diversity)~\cite{mtld_mccarthy2010} and VOCD (Vocabulary Diversity)~\cite{vocd-malvern2004} have been widely adopted, but they exhibit notable limitations. These popular measures often fail to account for critical factors such as the contextual relationship between tokens, the impact of duplication, and the variation in sentence length distributions across datasets \cite{vocd-length-problem,mtld_mccarthy2010}. Consequently, their applicability is constrained, particularly when addressing datasets with diverse linguistic and structural characteristics.

This problem is particularly challenging due to the inherent complexity of language. Datasets vary in vocabulary size, sentence structure, and contextual depth. While some approaches normalize vocabulary measures or incorporate sentence-length metrics \cite{DQI}, they often rely on oversimplified assumptions. These metrics fail to address key dimensions of dataset complexity, leading to potentially flawed evaluations in NLP tasks.

Our research addresses these gaps by focusing on vocabulary quality assessment through an autoencoder-based methodology. We demonstrate that reconstruction complexity can effectively measure vocabulary quality, providing insights into the underlying dataset characteristics. Through extensive validation, our framework establishes a robust approach for evaluating vocabulary quality in NLP datasets. Our contributions include: 
\begin{enumerate}
    \item A systematic critique of existing metrics, highlighting their limitations in real-world scenarios.
    \item An autoencoder-based framework that leverages neural network capacity as a proxy for vocabulary diversity and complexity.
    \item Evaluation of the framework using two datasets: DIFrauD~\cite{difraud} for deceptive text and Project Gutenberg~\footnote{https://huggingface.co/datasets/manu/project\_gutenberg} for diverse domains and languages.
\end{enumerate}

\section{Related Work}

\subsection{Lexical Metrics}

Lexical metrics like TTR, VOCD, and MTLD assess lexical diversity in linguistic datasets \cite{ttr_yule1944,vocd-malvern2004,mtld_mccarthy2010}. Each has strengths and limitations. TTR, simple and efficient, is unreliable for longer texts \cite{ttr-reliability-problem}. VOCD uses a probabilistic model but is complex and hard to interpret \cite{vocd-length-problem}. MTLD is robust to text length but introduces inconsistencies and lacks semantic considerations  \cite{mtld_mccarthy2010}.

Newer approaches like HD-D and MTLD+ address some shortcomings but still have limitations \cite{hd-d_jarvis2010,mtld_egbert2015}. These metrics overlook sentence-level variations, contextual relationships, and duplication impact. They treat all tokens equally, disregarding semantic and syntactic roles, and are sensitive to dataset size and domain-specific characteristics \cite{duplication_critique2021,mtld-vocd-problem}. This highlights the need for more robust, context-aware metrics in modern NLP applications.

\subsection{Vocabulary Quality in NLP}

Vocabulary quality is critical for dataset evaluation, directly influencing language model performance. Existing metrics like token counts or frequency distributions \cite{sharoff-2020-know} often neglect semantic richness, syntactic variability, and cross-domain generalizability. Normalizing vocabulary size by dataset size as in Data Quality Index 1 (DQI 1)~\cite{DQI} fails to account for sentence length differences, while duplication inflates scores without adding diversity. Research highlights biases in quality metrics due to oversampling or poor curation \cite{dodge2021}. A comprehensive taxonomy of NLP metrics, including vocabulary quality, diversity, and inter-class differences, is provided in \cite{dq_taxonomy}.

Cross-genre and multilingual datasets pose challenges due to distinct linguistic features \cite{lin2020} and language-specific variations \cite{conneau2020}. Recent advancements address these issues by evaluating contextual and structural properties. Tools assess domain diversity and lexical richness \cite{gururangan2020}, while token-level attention metrics measure vocabulary informativeness \cite{zhang2021}. Metrics like overlap percentage and token recall \cite{caswell2021} align vocabulary evaluation with downstream requirements, aiding domain-specific applications.

Despite these advancements, comprehensive vocabulary quality evaluation remains a challenge. Many metrics focus on surface-level characteristics, neglecting contextual relationships, semantic richness, and structural complexity. Our work proposes a novel autoencoder-based framework that evaluates vocabulary quality across multiple linguistic dimensions, providing insights for dataset development and analysis.

\section{Critique of DQI 1 and Other Vocabulary Metrics}
\begin{figure}[ht]
    \vspace{-10mm}
    \centering
    \includegraphics[width=0.8\textwidth]{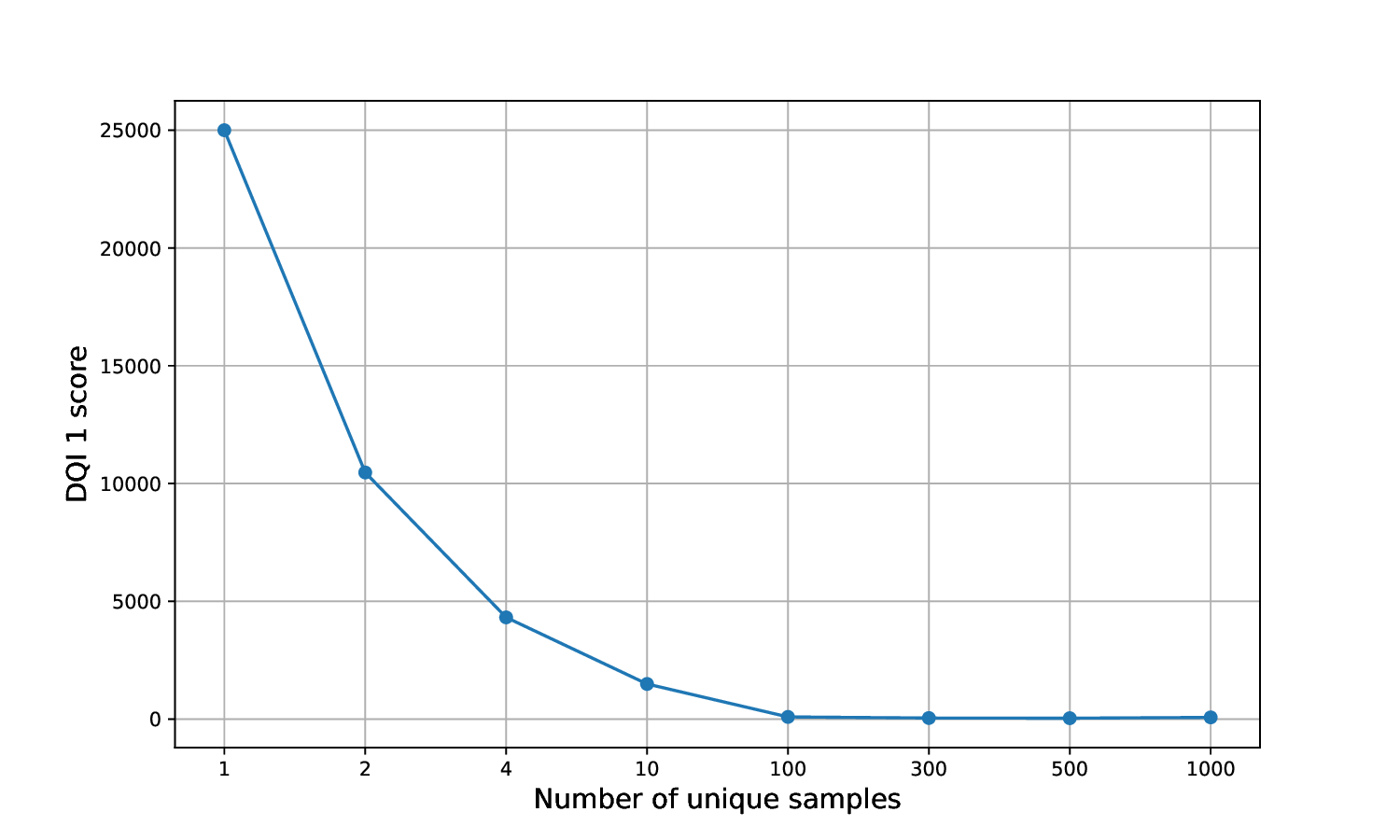}
    \caption{This plot shows the change in the DQI 1 metric (Y-axis), which measures vocabulary quality, across different numbers of unique rows in a fixed sample size of 1000 rows (X-axis). The DQI 1 value decreases as the number of unique rows increases.}
    \label{fig:dqi-duplication}
\end{figure}

Vocabulary quality is a critical aspect of evaluating NLP datasets. Several metrics have been proposed to measure vocabulary richness and diversity, including DQI 1, VOCD, and MTLD. While these metrics offer valuable insights, they exhibit notable limitations. This section provides a detailed critique of DQI 1 and compares it with the shortcomings of other metrics, such as TTR, VOCD, and MTLD.

\subsection{Critique of DQI 1}

The authors of \cite{DQI} propose DQI 1, which aims to measure dataset quality through a combination of two components: vocabulary and sentence length. However, the formulation of DQI 1 exhibits several critical issues, which limit its applicability and interpretability.

Let $X$ be the dataset, $v$ be the vocabulary, $s$ be the sentence length, $S$ be the set of all sentences in the dataset, $a$ and $b$ be the lower and upper thresholds of sentence length, and $size$ be the total number of samples, then the DQI 1 formula is as follows:
\begin{center}
    Vocabulary $= \frac{v(X)}{size(X)} + \sigma(s(X)) \cdot \frac{\sum_{S}(sgn((s-a)(b-s))}{size(S)}$
\end{center}

Firstly, the name ``Vocabulary'' in the DQI 1 metric is misleading. While the metric claims to focus solely on vocabulary, it incorporates sentence length into the formula, causing ambiguity. A more precise name, such as ``Vocabulary and Sentence-Length Composite,'' would better reflect its scope.

The vocabulary component oversimplifies richness by ignoring the saturation of unique tokens as datasets grow. Its normalization by dataset size assumes uniformity across domains, which is invalid for domains like social media with inherently shorter texts. Similarly, the sentence length component overlooks the impact of duplication, as normalizing by unique sentences distorts diversity representation. Figure~\ref{fig:dqi-duplication} demonstrates this flaw, showing that duplication paradoxically increases the DQI 1 score, contrary to expectations. While removing the sentence length component might seem like a solution, this would not address the fundamental issues with vocabulary normalization and would leave us with an incomplete measure of dataset quality.

\subsection{Critique of TTR, VOCD, and MTLD}

\begin{figure} [ht]
    \centering
    \includegraphics[width=1\linewidth]{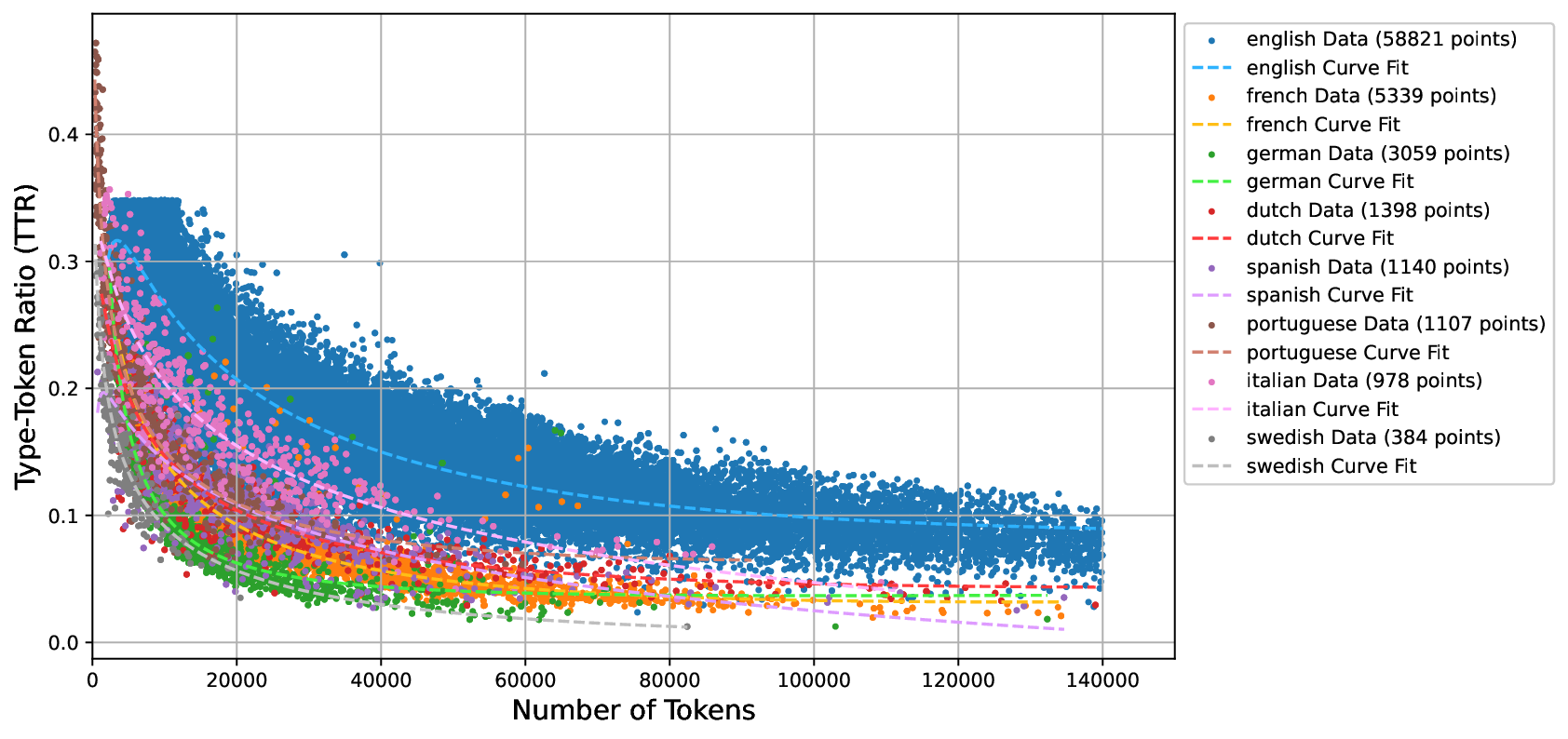}
    \caption{Scatter plot of Type-Token Ratios (TTR) vs. token counts for multilingual Gutenberg datasets, showing TTR's sensitivity to dataset size and its limitations for comparing vocabulary diversity in large corpora. Curve fits highlight consistent trends across languages.}
    \label{fig:TTR}
\end{figure}

\begin{figure} [ht]
    \centering
    \includegraphics[width=1\linewidth]{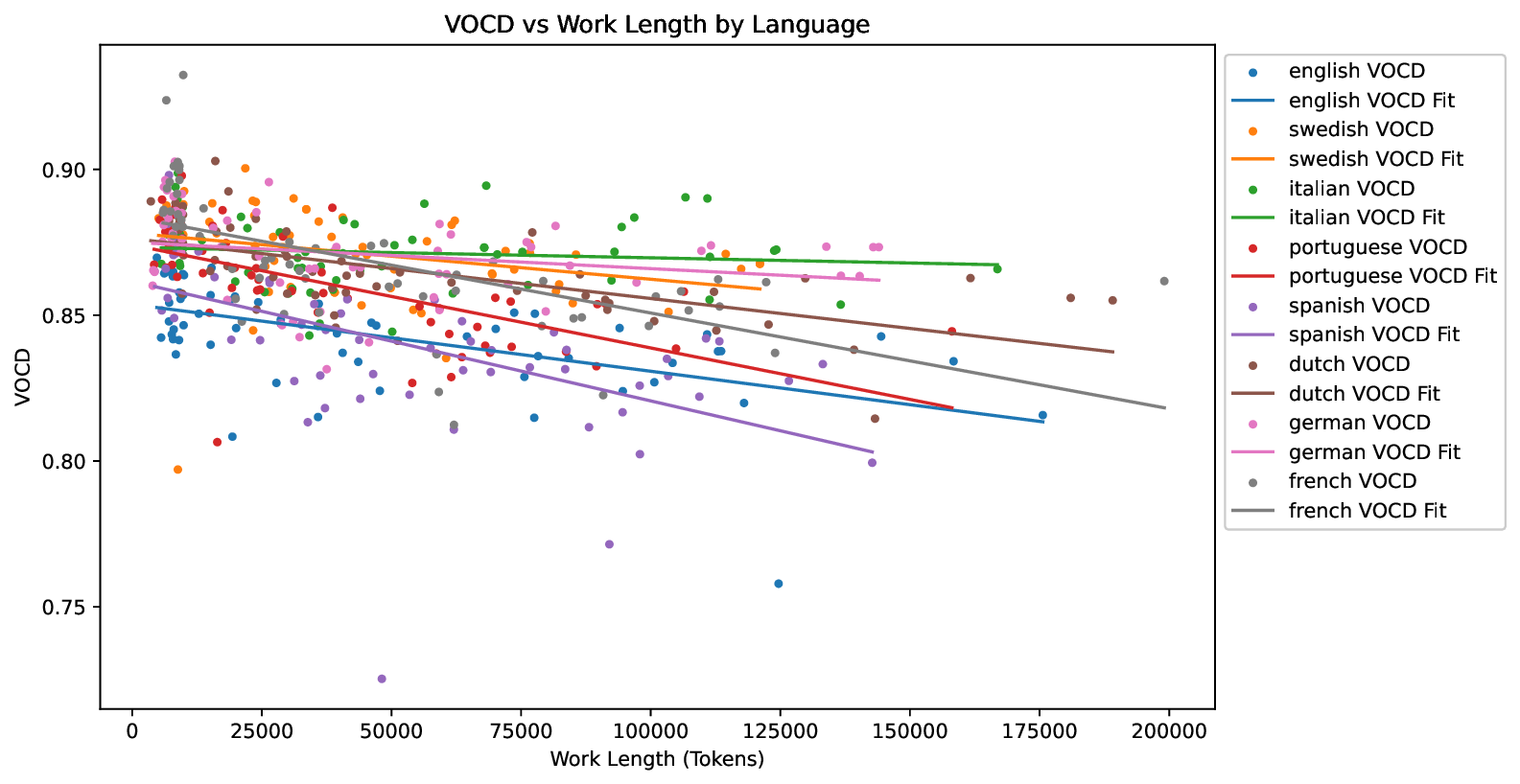}
    \caption{Scatter plot showing the relationship between VOCD and Work Length (in tokens) across multiple languages. Each color represents a different language, with corresponding linear fit lines indicating trends. The negative slope suggests a decline in VOCD as work length increases.}
    \label{fig:VOCD}
\end{figure}

\begin{figure} [ht]
    \centering
    \includegraphics[width=1\linewidth]{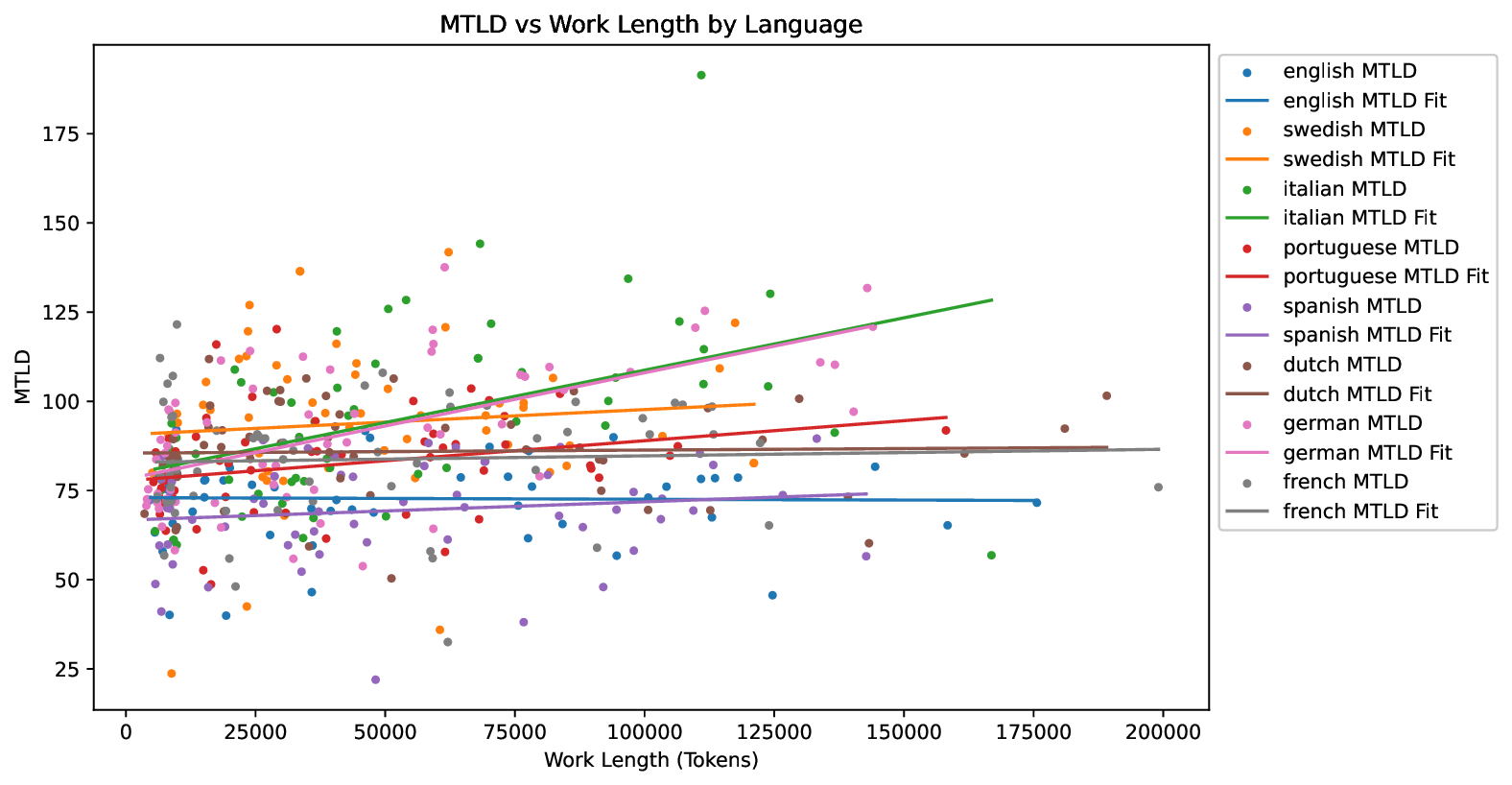}
    \caption{Scatter plot depicting the relationship between MTLD and Work Length (in tokens) across various languages. Each color represents a different language, with linear trend lines showing a generally positive correlation, suggesting that MTLD tends to increase slightly as work length grows.}
    \label{fig:MTLD}
\end{figure}

Metrics for lexical diversity, such as TTR, VOCD, and MTLD, are commonly used to evaluate dataset vocabulary quality. However, these metrics have limitations, especially for large-scale datasets.

TTR, the ratio of unique tokens to total tokens, is highly sensitive to dataset size, as shown in Figure~\ref{fig:TTR}. As datasets grow, TTR values decline rapidly, making it unreliable for evaluating vocabulary diversity at scale~\cite{ttr-reliability-problem}. Domain-specific biases further impact TTR: technical texts yield lower TTRs, while creative writing inflates them.

VOCD uses a probabilistic model to estimate lexical diversity, addressing some of TTR's limitations. However, it struggles with stability in short texts due to limited token counts affecting curve fitting and ignores deeper contextual relationships between words~\cite{vocd-length-problem}. Its lack of interpretability also reduces its practical utility. This limitation is evident in Figure~\ref{fig:VOCD}, which shows that VOCD values generally decline as work length increases across multiple languages. The downward slopes in the trend lines highlight VOCD’s sensitivity to text length, with longer texts experiencing a greater reduction in estimated vocabulary diversity. This suggests that while VOCD may capture broad lexical patterns, its effectiveness diminishes for extended works.

MTLD segments text, calculates TTR for each segment, and averages the results. This reduces sensitivity to dataset size, making it more reliable for cross-domain comparisons. However, MTLD is dependent on segment length choice and fails to capture deeper linguistic properties~\cite{mtld-segmentation-problem}. As demonstrated in Figure~\ref{fig:MTLD}, MTLD exhibits a slight positive correlation with work length across different languages, with trend lines indicating a more stable estimation of lexical diversity compared to VOCD. This suggests that MTLD provides a more consistent measure across varying text lengths, making it useful for cross-linguistic analyses. However, the variation in slopes across languages underscores its sensitivity to segment length configuration, which can still impact comparability.

In summary, current vocabulary metrics like DQI 1, TTR, VOCD, and MTLD have limitations in dataset size sensitivity, domain biases, and linguistic depth. To address these issues, we propose a neural network-based autoencoder methodology that evaluates vocabulary richness through model complexity, providing a more dynamic and comprehensive assessment framework.

\section{Methodology}
   
In this paper, we distinguish four key attributes of a dataset's vocabulary. First, vocabulary richness measures the raw count and range of unique words, indicating how extensive the vocabulary is. Second, vocabulary diversity examines how uniformly words are distributed across the text, revealing whether certain terms dominate or if usage is balanced. Third, vocabulary complexity analyzes surface-level linguistic patterns, including factors like sentence structure and word length. Finally, vocabulary quality evaluates the overall fitness and utility of the vocabulary for specific applications, considering factors like domain relevance and coverage. Our analysis focuses on intrinsic vocabulary characteristics, independent of pre-trained embeddings to avoid potential biases.

This section outlines our proposed methodology for evaluating these vocabulary attributes in NLP datasets. Addressing the limitations of metrics like DQI~1, TTR, VOCD, and MTLD, we propose an autoencoder-based approach that uses neural network complexity as a proxy for overall vocabulary quality -- particularly capturing facets of richness, diversity, and complexity. By analyzing autoencoder capacity requirements, our framework captures deeper insights into lexical characteristics, dynamically evaluating the interplay of vocabulary size, token distribution, and structural patterns without relying on static assumptions.

\begin{figure} [ht]
    \centering
    \includegraphics[width=1\textwidth]{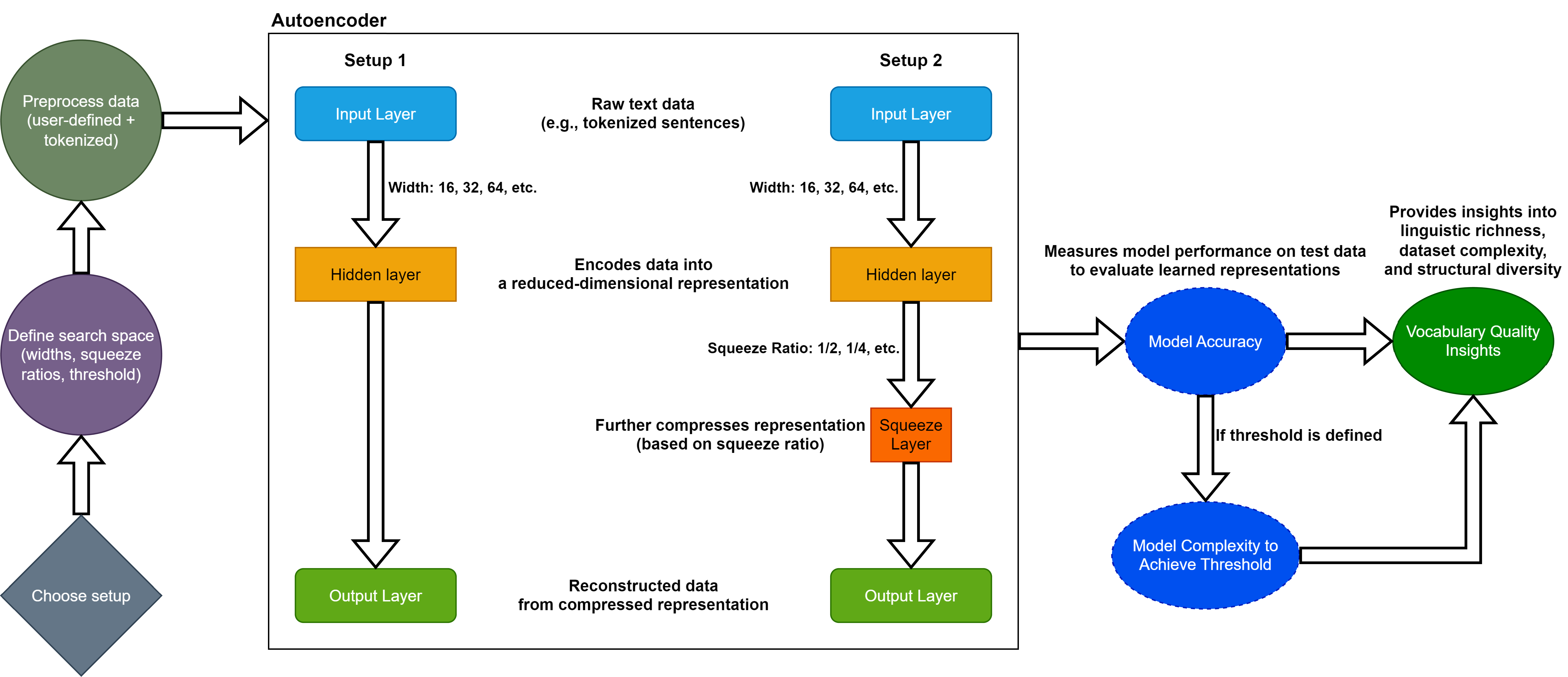}
    \caption{Framework for Evaluating Vocabulary Quality Using an Autoencoder
    Model. The process begins with preprocessing and setup definition, followed by autoencoder training and evaluation, resulting in model accuracy and insights into vocabulary richness, diversity, and complexity.}
    \label{fig:framework}
\end{figure}

\subsection{Autoencoder Framework}

As illustrated in Figure~\ref{fig:framework}, the proposed framework uses autoencoders to evaluate vocabulary quality. Autoencoders compress input data into lower-dimensional representations and reconstruct it, with the evaluation focusing on the model's capacity---the network size and configuration needed to achieve target accuracy. A dataset with a broad range of distinct tokens or more intricate token distributions (high richness, diversity, or complexity) will require a wider hidden layer to accurately reconstruct the text, whereas a repetitive dataset with fewer unique tokens can be reconstructed with a smaller network width, indicating lower diversity.

The input to the framework is any text, which undergoes preprocessing, including tokenization and user-defined transformations, as outlined in Figure~\ref{fig:framework}. The model aims to reconstruct the entire input text accurately. The output is the model's reconstruction accuracy given a specific configuration, including network widths and squeeze ratios. We selected a reconstruction accuracy threshold of 51\% for some experiments as a baseline to surpass random performance (50\%). While useful for consistent evaluation, this threshold is not always necessary, as trends in model performance and relative dataset complexity often provide sufficient insights without relying on fixed accuracy thresholds.

\subsubsection{Setup 1: Basic Non-bottlenecked Autoencoder}

In the first setup, we employ a non-bottlenecked autoencoder architecture with an input layer, a single hidden layer, and an output layer. The hidden layer width is varied across 64, 128, 256, and 512 neurons to assess its impact on reconstruction performance. ReLU activation~\cite{relu} is used. While we initially experimented with masking tokens from a Named Entity Recognition (NER) model to reduce bias, we ultimately decided against using this approach as it did not significantly impact relative performance.

We use the Keras Tokenizer~\footnote{https://www.tensorflow.org/api\_docs/python/tf/keras/preprocessing/text} for text preprocessing, focusing on surface-level linguistic features to evaluate vocabulary quality without biases from pretrained embeddings. The output of the tokenizer serves as the input to the autoencoder, configured with a learning rate of 0.001, trained for 100 epochs with a batch size of 32. The input dimension is set to the vocabulary size, and the output layer uses a softmax activation function. Reconstruction loss is computed using sparse categorical crossentropy, and the optimizer is Adam. Dataset complexity is inferred from the hidden layer width required to achieve a predefined reconstruction error threshold, with higher quality vocabularies necessitating wider layers for comparable performance.

\subsubsection{Setup 2: Squeezed Autoencoder}

To explore the relationship between compression and reconstruction, we add a squeeze layer between the hidden and output layers, forcing the model to condense information before reconstruction. The squeeze ratio, defined as the neurons in the squeeze layer relative to the hidden layer, is varied across 1/2, 1/4, 1/8, and 1/16 to evaluate how well the model compresses vocabulary and structural diversity.

This squeezed autoencoder provides a detailed assessment of dataset complexity, balancing compression with reconstruction fidelity. Richer vocabularies with intricate contextual relationships are expected to require larger squeeze ratios (closer to 1) to maintain accuracy, reflecting higher informational density.

\subsection{Rationale for Using Multi-Layer Perceptron (MLP)}

We employ Multi-Layer Perceptrons (MLPs)~\cite{mlp} as the foundational architecture for our autoencoders due to their simplicity, flexibility, and interpretability. MLPs are well-suited for systematically varying model capacity by adjusting layer widths and depths, making them an ideal choice for exploring the relationship between vocabulary quality and model complexity. Additionally, their straightforward architecture ensures that the results are not confounded by domain-specific nuances or task-specific optimizations inherent in more complex models such as transformers. By focusing on MLPs, we maintain a consistent baseline for evaluating datasets across different domains.

\subsection{Datasets}

We evaluate our methodology using two datasets: the DIFrauD dataset and the Project Gutenberg dataset, selected for their complementary linguistic and structural diversity. DIFrauD focuses on domain-specific challenges with five subsets: \textit{political statements}, \textit{job scams}, \textit{product reviews}, \textit{phishing}, and \textit{fake news}. These subsets represent a wide range of lexical patterns, such as rhetorical ambiguity in political statements, manipulative language in job scams and phishing, and linguistic differences in authentic versus fake reviews.

The Project Gutenberg dataset offers a multilingual benchmark, including English, Swedish, Italian, German, French, Portuguese, Spanish, and Dutch, while spanning diverse genres and historical periods. It is curated with a rigorous proofreading framework where multiple proofreaders meticulously verify each book page by page to ensure minimal OCR errors~\footnote{https://www.pgdp.net/c/}, making it an ideal dataset for evaluating vocabulary quality. Together, these datasets facilitate a comprehensive analysis of our framework across domain-specific, multilingual, and historically diverse corpora.

\section{Results and Discussions}
\subsection{DIFrauD Datasets - Setup 1}
\begin{figure}[ht]
    \vspace{-10mm}
    \centering
    \includegraphics[width=0.8\linewidth]{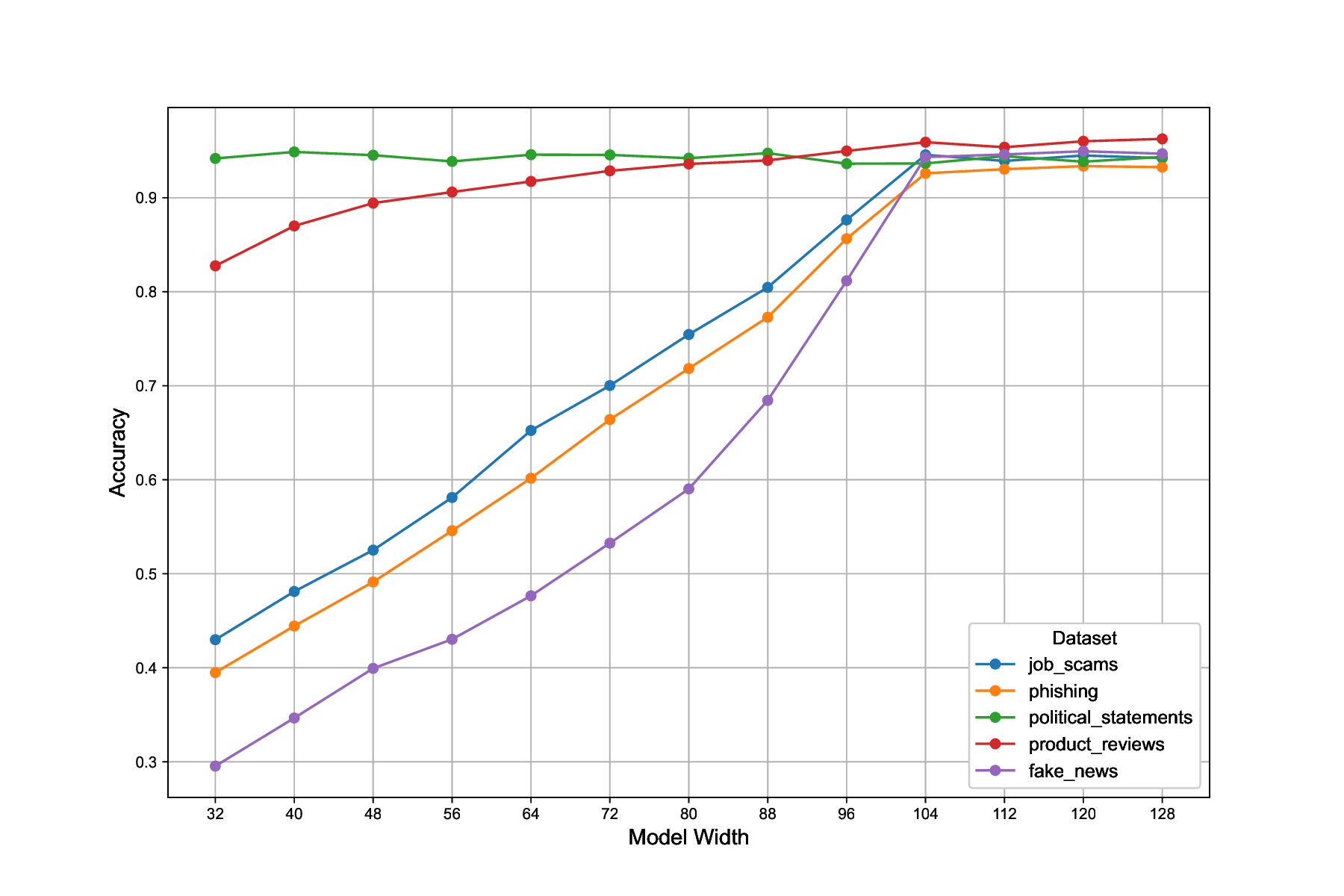}
    \caption{Accuracy Across Different Model Widths for DIFrauD Datasets: The plot highlights accuracy variations across configurations, showing how different datasets respond to model complexity, with some requiring greater width for optimal performance.}
    \label{fig:difraud}
\end{figure}

The results depicted in Figure \ref{fig:difraud} support our hypothesis that datasets with larger and more complex vocabularies require greater model width to achieve comparable performance to those of simpler datasets. Using a 51\% accuracy threshold, the \textit{political statements} and \textit{job scams} datasets meet this threshold at a width of 32, though \textit{job scams} exhibit lower performance and benefit more from increased widths, indicating greater vocabulary and structural complexity. In contrast, \textit{phishing} and \textit{fake news} datasets require higher widths, 56 and 72 respectively, to achieve the same threshold, reflecting their richer and more complex vocabularies. These findings highlight how datasets with complex linguistic properties demand greater model capacity for effective representation.

\subsection{Duplication Results}
\begin{figure}[ht]
    \vspace{-10mm}
    \centering
    \includegraphics[width=0.8\linewidth]{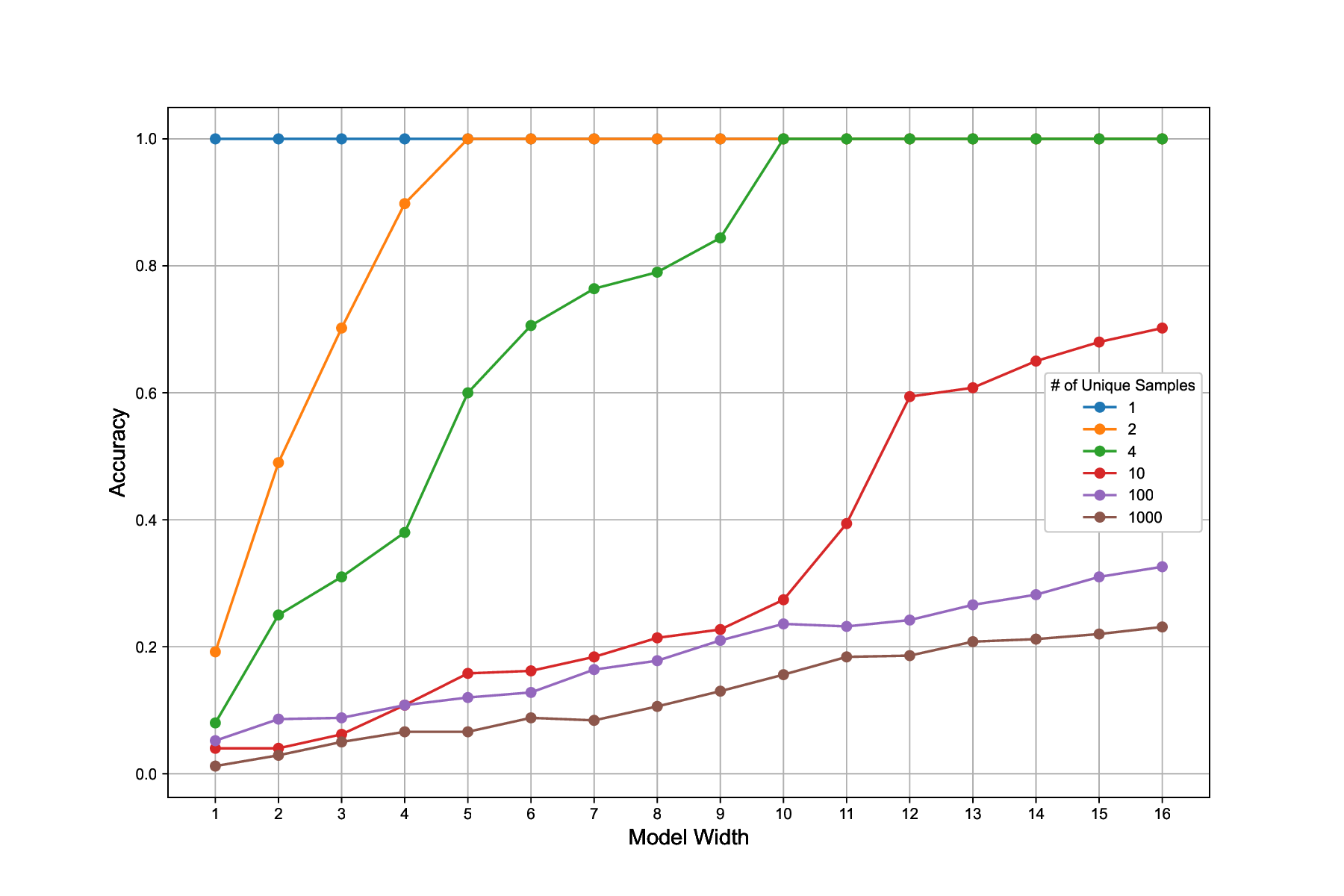}
    \caption{Impact of Model Width on Accuracy for Datasets with Varying Unique Sample Counts. Datasets with fewer unique samples achieve high accuracy with smaller widths, while richer vocabularies require wider models to capture greater linguistic diversity.}
    \label{fig:duplication_results}
\end{figure}

Figure~\ref{fig:duplication_results} illustrates that our approach maintains robust performance in assessing vocabulary characteristics, even when datasets contain many duplicate entries. For instance, datasets with only 1 or 2 unique samples reach high accuracy at minimal autoencoder widths, indicating limited linguistic diversity and lower overall vocabulary quality. This underscores the framework's capacity to recognize such datasets without demanding unnecessary model complexity.

Meanwhile, datasets featuring more unique samples (e.g., 10, 100, or 1000) require larger widths to achieve comparable accuracy, reflecting their richer and more varied vocabularies. Notably, datasets at intermediate levels of duplication (e.g., 10 or 100 unique samples) benefit from particular width configurations, highlighting the framework's flexibility in handling a range of complexities. In this way, our method accurately represents vocabulary characteristics while remaining resilient to duplication artifacts.

\subsection{Robustness Against Languages and Text Length}

\begin{figure}
    \centering
    \includegraphics[width=0.8\linewidth]{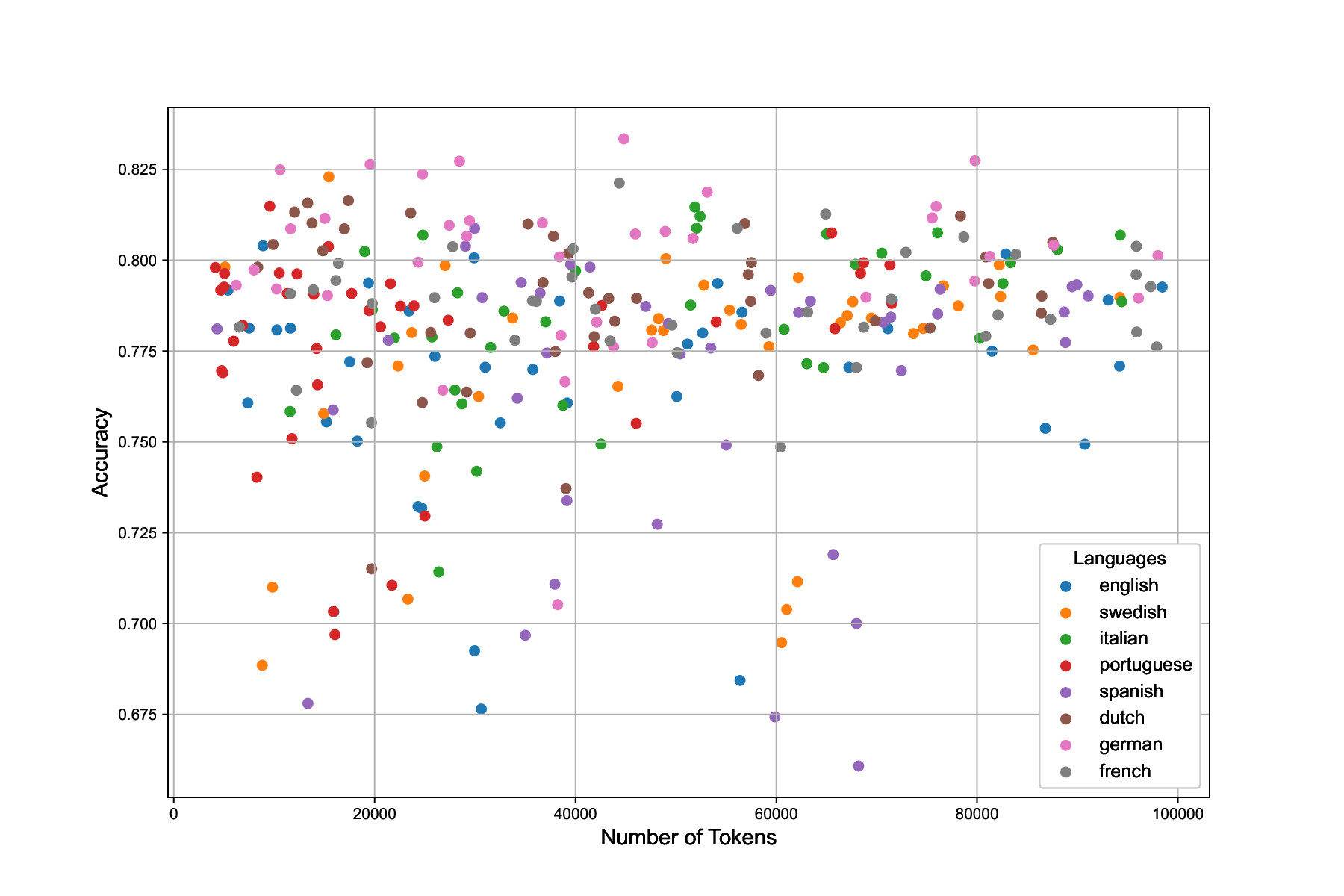}
    \caption{Accuracy vs Text Length across Languages. The figure demonstrates that performance is primarily influenced by vocabulary quality rather than text length or the language of the dataset. This is evidenced by the consistent distribution of accuracy values across languages, regardless of the text length.}
    \label{fig:multi_length}
\end{figure}

The results presented in Figure~\ref{fig:multi_length} reveal that the performance of our framework is primarily determined by vocabulary quality, as opposed to text length or the language of the dataset. Due to resource constraints, 40 works were randomly sampled from each language to ensure equal representation across the datasets. All evaluations were conducted with a fixed model width of 64, demonstrating consistent accuracy across languages and indicating that performance is independent of specific languages.

Statistical analyses further support this conclusion. A Pearson correlation analysis between accuracy and text length yielded a weak, non-significant correlation (r = 0.1219, p = 0.2814), suggesting text length has minimal influence on performance. Pairwise t-tests between languages revealed no significant differences, reinforcing that the framework evaluates datasets independently of linguistic features.

These findings demonstrate the robustness of our framework in assessing vocabulary characteristics across diverse datasets. By remaining unaffected by text length or language, the method provides a fair and reliable evaluation of vocabulary richness in multilingual contexts.

\subsection{18th-Century vs 20th-Century - Language Evolution}

\begin{table}[ht]
    \centering
    \small
    \renewcommand{\arraystretch}{1}
    \setlength{\tabcolsep}{6pt}
    \begin{tabularx}{\textwidth}{|c|l|X|X|X|X|} 
         \cline{3-6} 
         \multicolumn{2}{c|}{} & \multicolumn{4}{c|}{\textbf{Squeeze Ratio}} \\ \hline 
         \textbf{Century} & \textbf{Width} & \textbf{1/2} & \textbf{1/4} & \textbf{1/8} & \textbf{1/16} \\ \hline 
         18th & 64  & \textbf{0.801} & 0.731 & 0.608 & 0.406 \\ \hline 
         18th & 128 & \textbf{0.814} & 0.790 & 0.739 & 0.620 \\ \hline 
         18th & 256 & 0.803 & \textbf{0.808} & 0.790 & 0.745 \\ \hline 
         18th & 512 & 0.795 & 0.800 & \textbf{0.802} & 0.786 \\ \hline 
         20th & 64  & \textbf{0.855} & 0.831 & 0.783 & 0.663 \\ \hline 
         20th & 128 & \textbf{0.845} & 0.839 & 0.832 & 0.788 \\ \hline 
         20th & 256 & 0.836 & \textbf{0.840} & 0.836 & 0.831 \\ \hline 
         20th & 512 & 0.838 & \textbf{0.839} & 0.837 & 0.836 \\ \hline
    \end{tabularx}
    \caption{Impact of Model Width and Squeeze Ratio on Accuracy for 18th- and 20th-Century Datasets. Bold values indicate the highest accuracy for each row, reflecting optimal performance at specific width.}
    \label{tab:gutenberg_squeeze}
\end{table}

The results in Table~\ref{tab:gutenberg_squeeze} show that model width and squeeze ratio significantly impact accuracy for 18th-century and 20th-century datasets, each consisting of 50 works selected from Project Gutenberg and spanning multiple genres and authors. Narrower models (e.g., 64 neurons) exhibit sharp performance drops as the squeeze ratio decreases, particularly for 18th-century works, reflecting their richer vocabulary and complex linguistic structures. Wider models (256 and 512 neurons) demonstrate robustness to smaller squeeze ratios, effectively adapting to the increased complexity.

In contrast, the 20th-century dataset maintains consistently higher accuracy for all widths and squeeze ratios, suggesting simpler vocabularies and less intricate structures. This stability is supported by a pooled t-test comparing the 18th and 20th centuries across the configurations, which yields a t-statistic of -2.6609 and a p-value of 0.0081, confirming that the higher accuracy of the 20th-century dataset is statistically significant. This highlights their reduced modeling demands compared to 18th-century datasets.

\subsubsection{Results Summary}

Overall, these results demonstrate the framework's versatility in assessing vocabulary attributes across diverse datasets. In the DIFrauD experiments, datasets with more complex and diverse vocabularies demanded wider autoencoder configurations, while duplication tests underscored the model's sensitivity to linguistic diversity. Multilingual evaluations further confirmed its resilience across varying languages and text lengths. Lastly, the historical experiment showcased the method's adaptability: 18th-century texts required greater capacity and exhibited higher sensitivity to reduced squeeze ratios compared to 20th-century texts.

\section{Limitations}
While our methodology demonstrates the effectiveness of autoencoder-based approaches for assessing vocabulary richness, diversity, and complexity, several limitations merit attention. First, the computational overhead of training multiple autoencoder configurations can challenge real-time applications or resource-constrained settings. Additionally, although our approach successfully identifies structural and lexical intricacies, its performance in noisy or low-resource datasets remains underexplored, highlighting the need for further research to enhance robustness and generalizability across diverse real-world scenarios.

Our use of learnability---measured by how well an autoencoder reconstructs text---as a proxy for these vocabulary attributes assumes that data exhibiting richer, more diverse, and structurally coherent token distributions are inherently more predictable and thus more accurately reconstructed. Empirical evidence supports this premise, indicating that autoencoders yield higher reconstruction errors when faced with noisy or low-quality inputs~\cite{reconstruction_errors}. However, relying solely on reconstruction quality has clear drawbacks. It does not directly capture semantic context, label correctness, or broader communicative intent. Consequently, while unsupervised learnability offers valuable insights into lexical richness, distributional diversity, and structural complexity, it should be used alongside other metrics for a well-rounded evaluation of dataset quality.

\section{Conclusion and Future Works}

This paper presents a new autoencoder-based methodology for evaluating vocabulary quality in NLP datasets. Our framework treats model capacity as a stand-in for linguistic complexity, providing a flexible assessment that exceeds the capabilities of traditional metrics. Through extensive experiments, we show that richer vocabularies demand wider models, linguistic diversity has a pronounced effect on model behavior, and our method remains robust across multiple languages and text lengths. Notably, the historical analysis highlights that 18th-century texts require larger model capacities and exhibit greater sensitivity to reduced squeeze ratios than those from the 20th century, demonstrating the framework’s adaptability to varying linguistic complexities. Future work includes extending the method to noisy and low-resource datasets and incorporating contextual embeddings to enrich the evaluation.

\section*{Acknowledgments}

Research partly supported by NSF grants 2210198 and 2244279, and ARO grants W911NF-20-1-0254 and W911NF-23-1-0191. Verma is the founder of Everest Cyber Security and Analytics, Inc.

\bibliographystyle{splncs04}

\end{document}